\documentclass[letterpaper, 10 pt, conference]{ieeeconf}  

\IEEEoverridecommandlockouts                              

\usepackage{amsmath} 
\let\oldthebibliography\thebibliography
\let\endoldthebibliography\endthebibliography
\renewenvironment{thebibliography}[1]{%
  \oldthebibliography{#1}%
  \setlength{\itemsep}{-0.6pt}%
  \setlength{\parskip}{0pt}%
}{\endoldthebibliography}
\usepackage{algorithmic}
\usepackage{algorithm}
\usepackage{array}
\usepackage[caption=true,font=footnotesize,labelfont=sf,textfont=sf]{subfig}
\usepackage{textcomp}
\usepackage{stfloats}
\usepackage{url}
\usepackage{verbatim}
\usepackage{graphicx}
\usepackage{cite}

\usepackage{footnote}
\usepackage[bottom]{footmisc}
\usepackage{threeparttable}
\usepackage{multirow}
\usepackage{tabularx}
\usepackage{ragged2e}
\usepackage{booktabs}
\usepackage{makecell}
\usepackage{xspace}
\usepackage{diagbox}
\usepackage{xcolor}
\usepackage{soul}
\usepackage[final]{pdfpages}
\usepackage{hyperref}

\makeatother

\newcolumntype{C}[1]{>{\centering\arraybackslash}p{#1}}
\newcolumntype{P}[1]{>{\centering\arraybackslash}p{#1}}
\newcolumntype{M}[1]{>{\centering\arraybackslash}m{#1}}
\newcolumntype{L}[1]{>{\raggedright\arraybackslash}p{#1}}
\newcolumntype{R}[1]{>{\raggedleft\arraybackslash}p{#1}}
\newcolumntype{J}[1]{>{\justifying\arraybackslash}p{#1}}

\IEEEoverridecommandlockouts
\begin{document}

\title{\LARGE \bf Framework and Multi-modal Dataset for Roadwork Zone Detection and Geo-localization}

\author{Zhiran~Yan$^{1 \;*}$,
        Yutong~Xin$^{3}$,
        S~Shyam~Shenoi$^{1}$,
        Rui~Song$^{2,3}$,
        and Gordon~Elger$^{1,2}$
\thanks{This work was supported by the Federal Ministry for Economic Affairs and Climate Action in the project "Gaia-X 4 AMS".}
\thanks{$^{*}$
Corresponding author, email address: zhiran.yan@thi.de
    }
\thanks{$^{1}$
Institute of Innovative Mobility (IIMo), Technical University Ingolstadt of Applied Sciences, Ingolstadt, 85049, Germany. 
    }
\thanks{$^{2}$
Fraunhofer Institute for Transportation and Infrastructure Systems IVI, Ingolstadt, 85051, Germany.
        }%
\thanks{$^{3}$
Technical University of Munich, Garching, 85748, Germany.
   }%
}

\maketitle

\begin{abstract}
Autonomous vehicles often rely on high-definition (HD) maps for navigation; however, these maps are not frequently updated and often lack semi-static information, such as temporary roadwork zones, which can significantly alter the road network. This limitation underscores the urgent need for an accurate global position of roadwork zones. However, the absence of publicly available datasets for evaluating roadwork zone detection and geo-localization models has hindered the development of reliable autonomous driving systems. To address this challenge, we propose the Roadwork Zone Detection and Geo-localization (RZDG) dataset, which includes both simulated and real-world data, providing multimodal sensor inputs along with comprehensive annotations. The dataset supports multiple perception tasks, including image semantic segmentation, 3D object detection, and object geo-localization. In addition, we introduce a tracker-based roadwork zone detection and geo-localization (RZDG) pipeline, an extension of AB3DMOT, for accurate object geo-localization in roadwork zones. We benchmark our approach on the RZDG dataset, demonstrating its effectiveness in detecting roadwork zones and transforming object positions from the local coordinate system to the global coordinate system. A prediction is considered a true positive (TP) if its estimated position falls within one meter of the ground truth. Our experimental results show that our approach achieves high accuracy on both real and simulated data. Specifically, we report: Precision: 0.565 (real) / 0.615 (simulated)
Recall: 0.898 (real) / 0.809 (simulated)
F1-score: 0.597 (real) / 0.665 (simulated).
The RZDG dataset and code can be found at: \url{https://github.com/chrisyan/RZDG}.

\end{abstract}

\section{Introduction}

\begin{figure}[htbp] \centering {\includegraphics[width=\columnwidth]{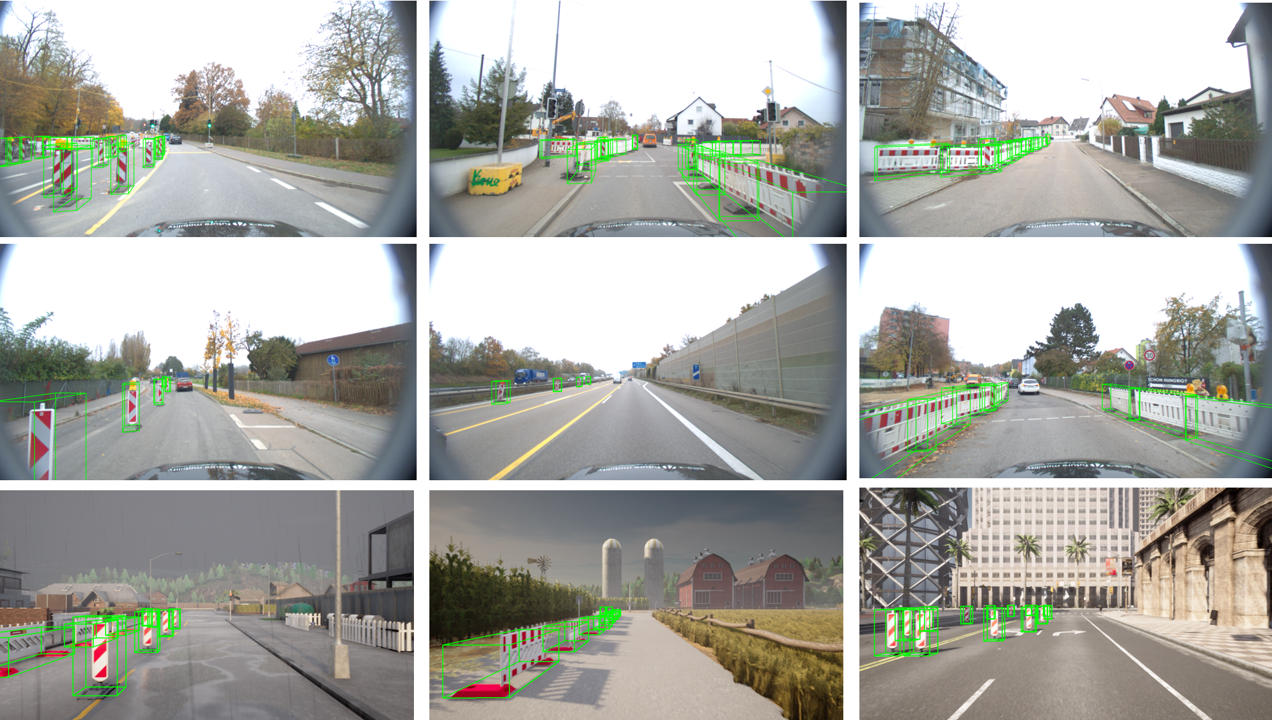}} \caption{Examples of diverse roadwork zones depicted in different urban contexts and various weather conditions, illustrating the complexity and crowded nature of these environments where objects such as barriers and road beacons often have indistinct boundaries. The first and second rows show real data from typical roadwork zones in Germany, while the third row presents simulated data from CARLA simulator.} \label{fig_dataset} \end{figure}

Efficient traffic routing in smart cities heavily relies on accurate and up-to-date high-definition (HD) maps. However, these maps often fail to capture temporary roadwork zones, which are critical for ensuring safe and efficient navigation. Roadwork zones disrupt daily traffic flow, pose significant safety hazards, and increase the risk of accidents, especially for autonomous vehicles (AVs) that depend on precise environmental perception. While HD maps provide detailed static information about road networks, they are not frequently updated to reflect transient changes such as roadwork zones. This limitation creates a critical gap in the situational awareness of AVs, potentially leading to unsafe driving behaviors or inefficient route planning.

Accurate and real-time detection and geo-localization of roadwork zones are essential for enhancing traffic safety, mitigating congestion, and optimizing construction efficiency. A Local Dynamic Map (LDM) serves as a database that integrates both static and dynamic data, providing a common reference system for AVs \cite{garcia2022ildm}. Notably, the geographic location and shape information of roadwork zones fall under the category of transient dynamic data \cite{etsi2009intelligent}. While previous studies have focused on detecting the presence of roadwork zones using camera-based systems \cite{Sundharam}, \cite{Oh}, and \cite{Katsamenis}, these methods are primarily designed for Advanced Driver Assistance Systems (ADAS) and do not provide sufficient information for trajectory prediction or global positioning, which are crucial for fully autonomous driving.

Recent advancements in roadwork zone detection involve using cameras or LiDAR to identify boundaries and create polygons from keypoints of 2D or 3D bounding boxes \cite{Shi}. However, these methods face significant challenges in real-world scenarios. For instance, roadwork zone boundaries are often unclear due to irregularly and sparsely distributed objects such as beacons, barriers, or traffic cones. This can lead to inaccuracies in generated polygons, such as overextended convex hulls that mislabel areas as roadwork zones when clear physical boundaries are absent. Moreover, these methods operate within the local coordinate system (e.g., the vehicle coordinate system), which does not effectively update the HD map with transient dynamic information required for the LDM.

To address these limitations, we propose a Roadwork Zone Detection and Geo-localization (RZDG) dataset, which combines both simulated and real-world data, providing multi-modal sensor data and comprehensive annotations. Our dataset supports multiple perception tasks, including image semantic segmentation, 3D object detection, and object geo-localization. Additionally, we develop a tracker-based Roadwork Zone Detection and Geo-localization (RZDG) pipeline, an extension of the AB3DMOT framework \cite{weng2020ab3dmot}, specifically designed for the precise geo-localization of roadwork zones. The primary objective of this pipeline is to accurately determine the global position of each detected object, thereby compensating for the transient information gaps in HD maps.

To summarize, our contributions are twofold: 
\begin{itemize} \item We construct the RZDG dataset, which includes both real-world and simulated data. The dataset features multi-modal sensor data (camera, LiDAR, and GPS/IMU) and comprehensive annotations such as semantic segmentation masks, 2D/3D bounding boxes, and global position coordinates of objects related to roadwork zones.

\item We develop a tracker-based RZDG pipeline that leverages camera, LiDAR, and GPS/IMU data to accurately determine the global positions of roadwork zones. This pipeline is easily extended to be capable of geo-localizing any trained class of objects depending on object detector, making it versatile for various applications. \end{itemize}

\section{Related Work}
\subsection{Roadwork Zone Detection Dataset}

Roadwork zones pose increased risks of traffic accidents \cite{Shahin} and can alter road networks. Detecting these zones helps alert drivers and self-driving vehicles, reducing accidents and improving navigation. Most studies focus on detecting work zone objects \cite{Sundharam}, \cite{Oh},\cite{Katsamenis}, while others define work zone detection as identifying polygons in the vehicle's Bird’s Eye View (BEV) \cite{Shi}. However, these methods are limited to the local coordinate system and fail to address the lack of temporary roadwork zones in HD maps. Table~\ref{tab:num_workzone} compares existing roadwork zone datasets with our proposed RZDG dataset.

\begin{table*}[ht]
\caption{Comparison of roadwork zone relevant datasets with our roadwork zone detection and geo-localization (RZDG) dataset}
\begin{center}
\begin{tabular}{p{1.25cm}p{1.25cm}p{1.25cm}p{1.25cm}p{1.25cm}p{1.25cm}p{1.25cm}p{1.25cm}p{1.25cm}p{1.25cm}}
\hline
\textbf{Datasets} & \textbf{Mapillary}  & \textbf{ApolloScape} & \textbf{BDD100K} & \textbf{nuScenes}& \textbf{WZDetection}  & \textbf{LowCost} & \textbf{ROADwork} & \textbf{RoSA}& \textbf{RZDG (ours)} \\
\hline
Year                        & 2017  & 2018 & 2020 & 2020& 2021 & 2023 & 2024& 2024& 2025 \\
\hline
Publicly$^{\mathrm{a}}$     & yes   & yes  & yes& yes& no& no& yes& no& yes \\
\hline
Real/sim                    & real  & real  & real  & real  & real  & real  & real  & real  & real/sim  \\
\hline
\# RGB images               & 25K   & 300K  & 100M & 1.4M & N/A& 3.3K & 7.4K & 2.6K         & 1.3K/8.5K\\
\hline
\# LiDAR points             & no    & 70K   & no  & 400K & N/A& no & no & no                & 1.3K/8.5K\\
\hline
\# GPS/IMU                  & no    & 140K  & N/A & N/A & no & no &  no$^{\mathrm{b}}$ &no  & 1.3K/8.5K\\
\hline
\# Seg. mask                & 25K   & 140K  & 7K  & no & no & no & 7416 & 2664              & 859/8.5K\\
\hline
\# 3D Boxes                 & no    & 70K   & no  & 1.4M & N/A & no & no & no               & 12K/67K\\
\hline
\# Object's global position & no    & no    & no  & no & no& no & no & no                   & 43/673\\
\hline
\# Class                    & 66    & 27 & 10 & 23 & 5& 1 & 15 & 1                          & 2/2\\
\hline
Location                    & World & China & USA & USA, Singapore & USA, Singapore  & USA, Korea, Canada &  USA&  Korea& Germany, CARLA\\
\hline
\multicolumn{10}{l}{$^{\mathrm{a}}$ At the time of publication of this paper.} \\
\multicolumn{10}{l}{$^{\mathrm{b}}$ Only images are geo-tagged.} \\
\end{tabular}
\label{tab:num_workzone}
\end{center}
\end{table*}

\subsection{Image Semantic Segmentation}
In recent years, significant progress has been made in 2D semantic segmentation algorithms based on deep learning \cite{deeplabv3p}, \cite{swin} and \cite{segformer} to classify each pixel in an image, which can provide detailed information such as the exact shape and boundaries of an object. However, it does not provide depth information of the object. 

\subsection{3D Object Detection}
Depending on the sensors, 3D object detection methods can be categorized into three types: LiDAR-based, Camera-based, and fusion-based detection.
\textbf{Camera-Based}:
\cite{li2019gs3d}, \cite{cai2020monocular} build upon 2D bounding boxes for 3D object detection.
\cite{Mono3d}, \cite{liu2021ground} are based on 3D anchor generation, producing 3D object proposals by assuming a ground plane prior and employing ground-aware convolution.
\cite{SMOKE}, \cite{MonoFlex} obtains depth information through keypoint estimation, using the keypoints of the object to infer its spatial positioning in the 3D space.
\textbf{LiDAR-Based}:
Methods such as \cite{PointNet}, \cite{DGCNN}, \cite{KPConv} process raw point clouds directly, extract feature vectors for each point and group them into high-dimensional representations. 
\cite{VoxelNet}, \cite{SECOND}, \cite{PV-RCNN} break the continuous point cloud into discrete 3D voxel grids to generate predictions.
\cite{PointPillars} introduces an optimization to the traditional 3D voxel grid by converting the 3D space into a columnar grid structure("pillars"). The encoding of point cloud data is then performed using 2D convolutional layers, which enhances the speed of the feature encoder while maintaining competitive accuracy.
\textbf{Fusion-Based}:
LiDAR-based detections often lack detailed texture information, whereas camera-based detections face challenges related to nighttime use and depth information loss\cite{limits}.
\cite{MVXNet} proposed two multi-modal fusion methods, PointFusion and VoxelFusion, to combine RGB image features with point cloud features. \cite{3DDualFusion} designed a Dual-Fusion Transformer that enhances detection performance by converting and fusing camera domain features with voxel domain features.

\subsection{Object Geo-localization}
Geo-localization refers to the process of determining an entity's geographical location, typically in the form of Global Positioning System (GPS) coordinates \cite{Wilson_Zhang}.  Object geo-localization techniques are used in a wide range of applications, such as locating potholes, cracks, and ruts in road maintenance applications \cite{govada2020road}, locating static objects such as traffic signs and traffic lights in autonomous driving applications \cite{Wilson}, \cite{Chaabane} and \cite{Krylov}, and locating dynamic objects such as target traffic users \cite{Namazi} for vehicle-to-everything (V2X) communication. However, none of the studies addressed the representation of the roadwork zone in the global coordinate system. The most decisive challenge in the task of object geo-localization is to develop an algorithm that combines the repeated detections of same object across multiple frames into a single prediction. The algorithm needs to identify these recurring objects and integrate their positional information, pose, and other data to generate a unique and accurate geo-localization prediction for each actually present object. To address this gap, we contribute the first roadwork zone detection and geo-localization benchmark with a multi-sensor dataset.
\section{Roadwork Zone Detection and Geo-localization (RZDG-real) Dataset}
We construct Roadwork Zone Detection and Geo-localization (RZDG) dataset, which consists of both real (RZDG-real) and simulated data (RZDG-sim), featuring multi-modal sensor data and comprehensive annotations. The annotations include image semantic segmentation masks, 3D bounding boxes, and the global positions of objects. To facilitate accessibility and utility for researchers, we structure the dataset in the KITTI format \cite{kitti}. The dataset supports tasks such as image semantic segmentation, 3D object detection, and geo-localization of roadwork zones.

\subsection{Data Acquisition} 
\begin{figure}[t!]
   \centering
   \includegraphics[width=0.8\columnwidth]{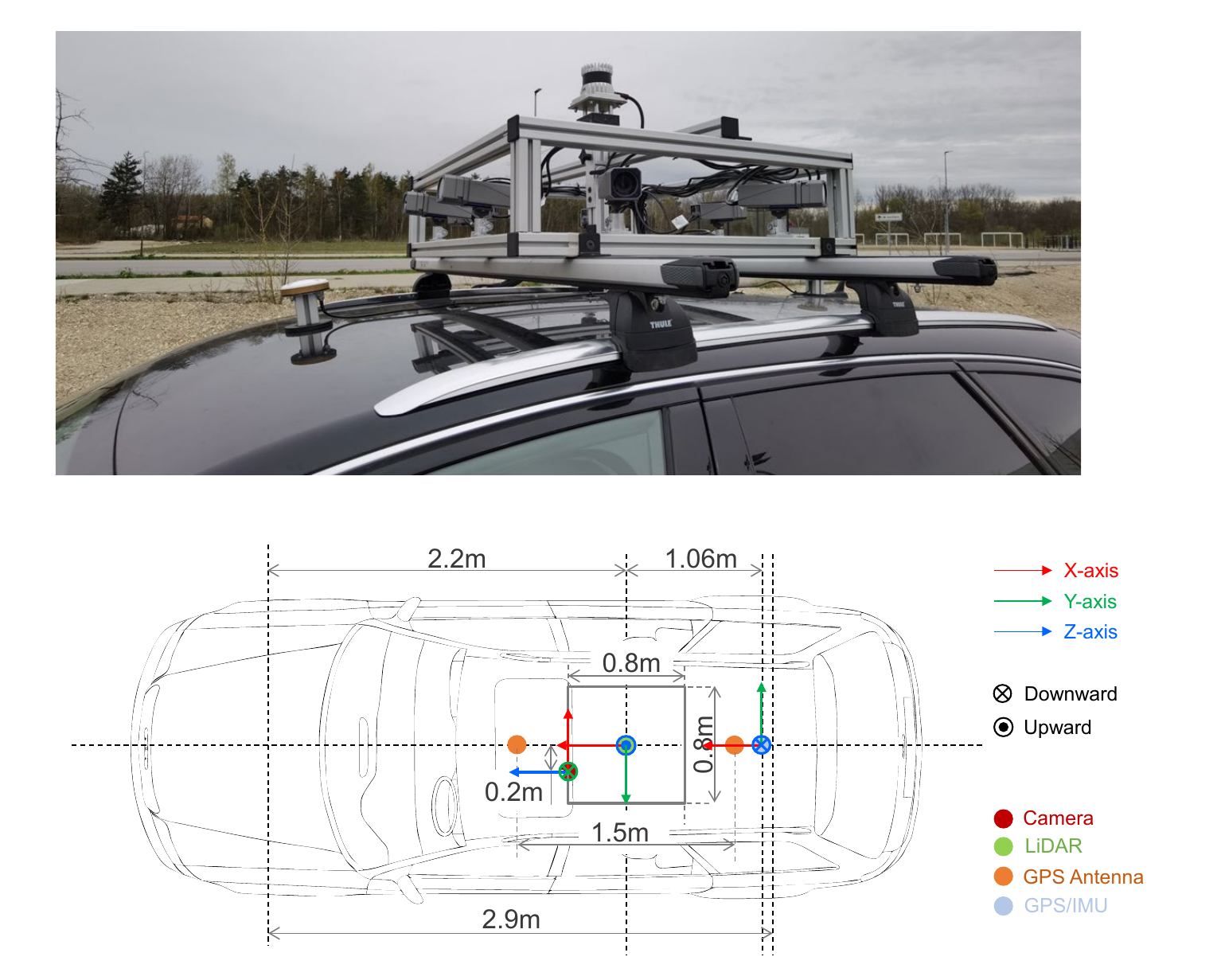}
   \caption{\textbf{Data Collection Platform.} This is an Audi A6\cite{bibliocad_audi_a6} equipped with seven Basler cameras, one Ouster OS1-128 LiDAR, three ARS548 RADARs and one Certus GPS/IMU sensor. (Note that the dataset in this work contains only one camera, one LiDAR, and one GPS/IMU.)}
   \label{fig:real_sensor_setup}
\end{figure}

\textbf{Data Collection.} Our data collection utilizes an Audi A6 outfitted with a Basler camera, an Ouster LiDAR, and a Certus GPS/IMU as shown in Fig.~\ref{fig:real_sensor_setup}. The detailed parameters are listed in Table ~\ref{tab:real_sensor_setup}. The vehicle’s onboard system runs on Ubuntu 20.04 and utilizes the ROS2 Foxy \cite{macenski2022robot} framework. We select 28 representative scenarios, each lasting between 10 to 100 seconds, which depict the vehicle navigating through various roadwork zones. These scenarios include diverse objects like road beacons and barriers, captured under varying conditions.

\begin{table}[h]
\caption{Sensor Specifications in real sensor car}
\begin{center}
\begin{tabular}{p{2.0cm}|p{5.5cm}}
\hline
\textbf{Sensors} & \textbf{Details} \\
\hline
1x camera &   Basler ace2R a2A1920-51gcBAS, with 4 mm lens \\
1x LiDAR & Ouster OS1-128, 128 vert. layers, 360°
HFOV, -22.5° to 22.5° VFOV, 120m capturing range   \\
1x GPS/IMU & Certus, a dual antenna GNSS/INS, 0.01m horizontal position accuracy with RTK, 0.015m vertical position accuracy with RTK, 0.1° roll, pitch, heading accuracy    \\
\hline
\end{tabular}
\label{tab:real_sensor_setup}
\end{center}
\end{table}

\begin{figure}[htbp]
\centering
{\includegraphics[width=0.85\columnwidth]{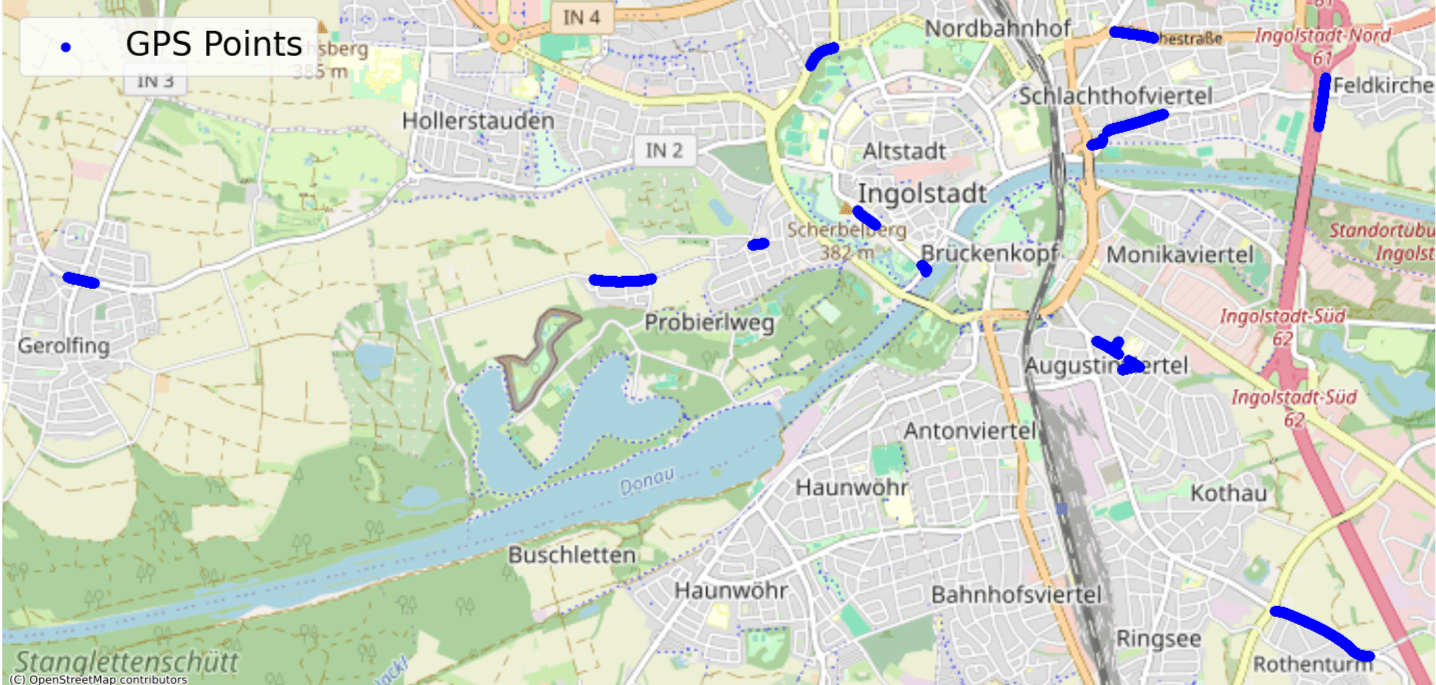}}
\caption{\textbf{Driving Route.} Driving routes are depicted using the sensor car's GPS data on OpenStreetMap, with the paths highlighted in blue to represent the car's trajectory.}
\label{fig_driving_route}
\end{figure}

\textbf{Driving Route.}
We gather data on various roadwork zones in Ingolstadt, Germany, under diverse conditions including different weather scenarios and at various times of the day. In Fig.~\ref{fig_driving_route}, we illustrate the sensor car's driving routes using GPS data displayed on OpenStreetMap \cite{OpenStreetMap}, providing a clear visual representation of the areas covered.

\textbf{Time Synchronization and Calibration.} Our computational system is precisely synchronized with a PTP server to ensure uniform system time across all sensors. Camera is hardware-triggered with LiDAR's 20 Hz pulse signal and operate at a frequency of 20 Hz. This setup ensures tightly aligned temporal data acquisition. GPS and IMU data are captured at a higher frequency of 40 Hz to capture more granular movement details. For calibration, we utilize Zhang's method \cite{zhang2000flexible} to calculate the camera's intrinsic matrix. The extrinsic matrix between the LiDAR and camera coordinate systems is determined using the CalibAnything approach \cite{luo2023calib} ensuring precise alignment and integration of sensor data.


\textbf{Data Privacy.} Due to data protection, we anonymize license plates and heads of pedestrians on all RGB images, which is achieved by utilizing a fine-tuning YOLOv8 \cite{10533619}, an object detection algorithm.


\subsection{Data Annotation} 
\textbf{3D Bounding Boxes Annotation}. For the collected LiDAR data, we sample key frames at a rate of 2 Hz and annotate 3D bounding boxes using SUSTech-Points \cite{li2020sustech}, a robust open-source labeling tool. Common object classes associated with roadwork zones in Germany include barriers and road beacons, which are typically irregular in shape and small in size. Each object is annotated with a 7-degree-of-freedom 3D bounding box, which includes x, y, z coordinates for the centroid position, l,w,h dimensions for the bounding box extent, and yaw for the rotation angle. Both types of objects are stationary. To facilitate evaluation, we convert the format of the manually labeled data to conform with the KITTI format.

\textbf{Image Semantic Segmentation Annotation}. Similarly, we sample key frames at 2 Hz for the collected camera data and annotate them with semantic segmentation masks using LabelMe \cite{5483185}. The object classes are consistent with those used for the 3D bounding boxes.

\textbf{Object's Global Position Annotation}. To accurately determine the ground-truth location of each object in the global coordinate system (i.e., latitude and longitude), we use a GPS with RTK corrections that provide two-centimeter accuracy. We select 3 different scenes and position the GPS directly over the center of each object to capture the precise ground-truth values.

\subsection{Data Statistics} A total of 1357 frames of the point cloud data are annotated with 3D bounding boxes, capturing 8837 barriers and 3643 road beacons. On average, each frame contains 6.51 barriers and 2.68 road beacons. For semantic segmentation, 859 frames are annotated, with barrier accounting for approximately 70M pixels and road beacon for 10M pixels. The average number of pixels per frame is 81K and 11.6K for barrier and road beacon, respectively. In addition, we accurately obtain the ground-truth global position for 43 objects across three representative roadwork zones.

\section{Roadwork Zone Detection and Geo-localization (RZDG-Sim) Dataset}
To minimize the costs associated with human annotation and to investigate the viability of simulated datasets for real-world applications, we develop the RZDG-Sim dataset using the CARLA simulator \cite{dosovitskiy2017carla}. This dataset provides multi-modal sensor data alongside annotations, including semantic segmentation masks, 3D bounding boxes, and the global positions of objects.

\subsection{Data Acquisition} The CARLA simulator is utilized to create a 3D environment representative of different roadwork zones, equipped with tools to automate data collection. Following the methodology outlined by \cite{schonborn2022rsa}, we incorporate 3D CAD models of typical German roadwork zone elements, such as barriers and road beacons. These objects are arranged on roads in configurations compliant with German traffic regulations, thereby forming roadwork zones of varying geometries and complexities.

For sensor data collection, the sensor suite on the ego vehicle is similar to the real sensor car, as shown in Fig.~\ref{fig:real_sensor_setup}. It includes an RGB camera, a 128-channel LiDAR, a GPS unit, and an IMU, all of which have been subjected to realistic noise augmentation to approximate true sensor performance. Each sensor is time-synchronized and calibrated to ensure data integrity. Data collection is conducted at a frequency of 10 Hz. Detailed specifications of the sensor setup are presented in Table ~\ref{tab:sim_sensor_setup}.

\begin{table}[h]
\caption{Sensor specifications in simulation}
\begin{center}
\begin{tabular}{p{2.0cm}|p{5.5cm}}
\hline
\textbf{Sensors} & \textbf{Details} \\
\hline
1x Camera & RGB, 1280*720, 90° horizontal FOV   \\
1x LiDAR & spinning, 128 channels, 2.6M points per second, 120 m capturaing range, -22.5° to 22.5° vertical FOV, 360° horizontal FOV   \\
1x GPS/IMU & 20 mm positional error, 2° heading error   \\
\hline
\end{tabular}
\label{tab:sim_sensor_setup}
\end{center}
\end{table}

\subsection{Data Annotation} For semantic segmentation masks, we utilize the built-in semantic segmentation camera provided by CARLA. For the 3D bounding boxes, we collect these only for objects within a 100-meter range of the camera's field of view (FOV), converting them into the KITTI format. Regarding global coordinates, we initially acquire each object's coordinates in the CARLA system and subsequently convert them to the WGS84 geodetic system.

\subsection{Data Statistics} Totally, 419 road beacons and 254 barriers are imported into 8 different towns to form roadwork zones in different types of roadways. The RZDG-Sim dataset contains 8591 frames of sensor data and annotation data, consist of 203 distinct scenarios of roadwork zones across 8 towns, under varying weather and lighting conditions. Each scenario is defined by the data captured as the ego-vehicle navigates through a specific roadwork zone, under particular environmental settings. The annotations for this dataset include 2D semantic segmentation masks, 3D bounding boxes in KITTI format, and the precise geographic locations of roadwork zone elements, e.g. road beacons and barriers. Overall, the dataset include 39,243 road beacons and 28,196 barriers, averaging approximately 4.57 road beacons and 3.28 barriers per frame.

\section{Roadwork Zone Detection and Geo-localization (RZDG) pipeline}

\begin{figure*}[t!]
   \centering
   \includegraphics[width=\textwidth]{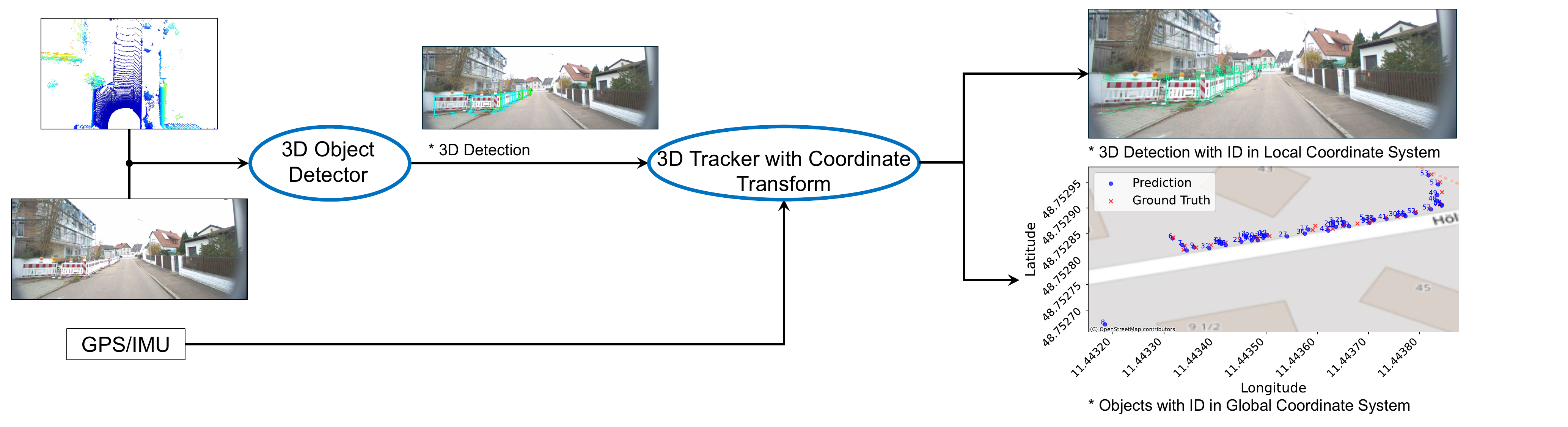}
   \caption{Overview of the Roadwork Zone Detection and Geo-localization (RZDG) Pipeline. This comprehensive pipeline is composed of three main modules: Initially, the 3D object detector captures 3D detections from LiDAR point cloud and/or camera image. Subsequently, these detections are processed by the tracker to form tracklets for each object detected across multiple frames. Finally, the tracklets are fused to create a unique detection for each object. This unique detection is transformed from the local coordinate system (i.e. camera coordinate system) to the global coordinate system using ego vehicle's GPS/IMU data, thereby enabling precise global positioning of each object.}
   \label{fig:overview}
\end{figure*}

The objective of object geo-localization is to accurately determine the global position of each object. We present a tracker-based Roadwork Zone Detection and Geo-localization (RZDG) pipeline, an extension of AB3DMOT framework \cite{weng2020ab3dmot}, specifically designed for the geo-localization of roadwork zones, as illustrated in Fig.~\ref{fig:overview}. The pipeline initiates with acquiring 3D detections from LiDAR point cloud and/or camera image. These detections are then processed through a tracking module to form tracklets for each detected object. Each tracklet is fused into a unique detection per object, which is subsequently transformed from the local coordinate system (i.e., the camera's coordinate system) to the global coordinate system using GPS/IMU data, ensuring precise prediction of the objects' global positions.

\subsection{3D Object Detection} As AB3DMOT is inherently a 3D tracker, it can incorporate 3D detections from any object detector that provides 3D bounding boxes. We evaluate the effectiveness of our proposed pipeline using three types of 3D object detectors: camera-based, LiDAR-based, and fusion-based detectors.

\subsection{Tracker-based Object Geo-localization} In the tracking module, 3D object detections from consecutive frames are fed into a tracker that utilizes a 3D Kalman filter with a constant velocity model for state estimation, combined with the Hungarian algorithm for data association. This process results in a series of tracklets, each representing the same object across multiple frames.

\begin{figure}[t!]
   \centering
   \includegraphics[width=0.15\textwidth]{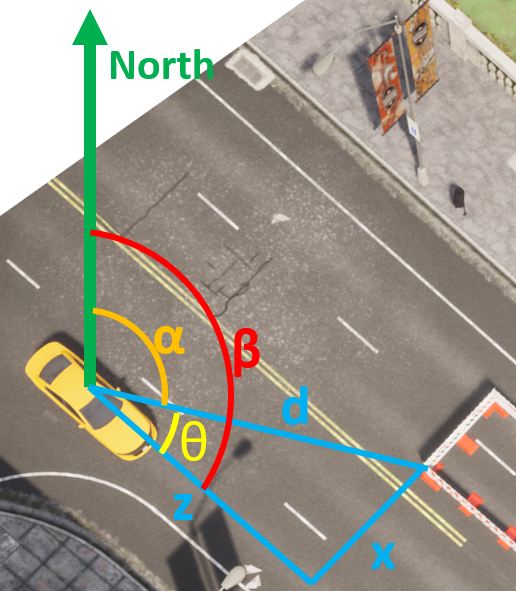}
   \caption{Required parameters for the coordinate transform module}
   \label{fig:geo}
\end{figure}

In the coordinate transform module, these tracklets are condensed into single object predictions represented by 3D bounding box with ID. The 3D bounding box's center point is then converted from the local coordinate system to the global coordinate system by leveraging the ego vehicle's GPS/IMU data. Fig. ~\ref{fig:geo} depicts the required parameters used in estimating an object’s global position. To accurately determine the global location of each object, we implement equations (1)-(7) from \cite{heidenreich1976distance}. Two strategies are used to fuse the tracklets for creating a unique detection for each object: 1) using the position from the last frame of the tracklet (LF) and 2) calculating a weighted average of the positions within the tracklet (WA), i.e. calculating the center of each tracklet and averaging the GPS/IMU data for frames in that tracklet. These strategies can effectively eliminate the repeated detection of the same object in different frames, thus obtaining a unique global position for each object.

\begin{equation}
d = \sqrt{z^2 + x^2}
\end{equation}
\begin{equation}
\theta = \arctan\left(\frac{|x|}{|z|}\right)
\end{equation}
\begin{equation}
\alpha = \beta - \theta
\end{equation}
\begin{equation}
lat_2 = \arcsin(\sin(lat_1) \cdot \cos(\frac{d}{R}) + \cos(lat_1) \cdot \sin(\frac{d}{R}) \cdot \cos(\alpha))
\end{equation}
\begin{equation}
U = \sin(\alpha) \cdot \sin(\frac{d}{R}) \cdot \cos(lat_1)
\end{equation}
\begin{equation}
V = \cos(\frac{d}{R}) - \sin(lat_1) \cdot \sin(lat_2)
\end{equation}
\begin{equation}
lon_2 = lon_1 + \operatorname{atan2}(U, V)
\end{equation}

 The object's location in the local coordinate system, i.e., the camera coordinate system, is given by \( z \) meters and \( x \) meters. The latitude and longitude of the ego vehicle are represented by \( lat_1 \) and \( lon_1 \), respectively, while \( lat_2 \) and \( lon_2 \) denote the latitude and longitude of the target object. The radius of the Earth, \( R \), is taken to be 6,372,800 meters. The variable \( d \) stands for the estimated distance between the camera on the ego vehicle and the target object. The angle \( \alpha \) is the estimated angle between the true North and \( d \), and \( \beta \) is the heading angle of the ego vehicle.
\section{Experiments}
To ensure a comprehensive and unbiased evaluation of the pipeline, we conduct separate assessments of the 3D object detection module and the geo-localization module. These evaluations are carried out on RZDG-Real and RZDG-Sim.

\subsection{3D Object Detection}
\textbf{Experiment Details}. We evaluate three 3D object detectors: SMOKE \cite{SMOKE}, Pointpillars \cite{PointPillars}, and MVXNet \cite{MVXNet}, on the two datasets, RZDG-Real and RZDG-Sim. The RZDG-Real dataset comprises 1085 training and 272 validation samples, split at a ratio of 0.8:0.2. The RZDG-Sim dataset includes 6872 training and 1719 validation samples, split at ratios of 0.8:0.2, respectively. Experiments are conducted on a platform with dual NVIDIA RTX 4090 GPUs. All detectors are implemented in mmDetection3D \cite{mmdet3d2020}, an open-source object detection toolbox based on PyTorch. The hyperparameters for training are following:  SMOKE  (batch size 8, learning rate 0.00025, 72 epochs), PointPillars  (batch size 48, learning rate 0.001, 80 epochs), and MVXNet  (batch size 2, learning rate 0.0001, 40 epochs).

\textbf{Evaluation}. Performance is assessed by calculating the Average Precision (AP) across three categories defined by varying levels of difficulty (easy, moderate, hard), following the KITTI criteria, which include distance, pixel size, occlusion, and truncation. Evaluations are conducted using an Intersection over Union (IoU) threshold of 0.5. Table \ref{tab:detection_results} details the performance of the three detectors in each category, while Table \ref{tab:inference_times} outlines the inference time for each detector on a single frame, utilizing the NVIDIA RTX 4090 GPU.

\begin{table}[ht]
\caption{Inference Time on RZDG-Real and RZDG-Sim Data. Results are shown with the best in each group in bold.\label{tab:inference_times}}
\begin{center}
\begin{tabular}{lcc}
\hline
Detector & RZDG-Real (ms) & RZDG-Sim (ms) \\
\hline
SMOKE          & 110.9 & 90.9 \\
PointPillars   & \textbf{102.0} & \textbf{85.9} \\
MVXNet         & 380.7 & 304.9 \\
\hline
\end{tabular}
\end{center}
\end{table}

\begin{table*}[ht]
\caption{3D Object Detection Results on RZDG Dataset (C=Camera, L=LiDAR). Results are shown with the best in each group in bold.\label{tab:detection_results}}
\begin{center}
\begin{tabular}{p{1cm}|p{1.5cm}p{1.25cm}p{1.5cm}|p{1.25cm}p{1.25cm}p{1.25cm}|p{1.25cm}p{1.25cm}p{1.25cm}}
\hline
& & & & & $\mathrm{AP}_{3D}$ (\%) & & & $\mathrm{AP}_{BEV}$ (\%) & \\
\cline{5-10}
 &Methods& Modality & Class& Easy & Moderate& Hard & Easy & Moderate& Hard \\
\hline
   Real&SMOKE& C & & 0.2830 & 0.2562 & 0.2562 & 0.6722 & 0.6129 & 0.6129 \\
  & PointPillars& L  & Road beacon  & 32.4564 & 30.3099 & 30.3099  & 53.9370 & 51.0171 & 51.0171    \\
  &MVXNet& C, L &  & \textbf{39.2720} & \textbf{38.6070} & \textbf{38.6070} & \textbf{58.3262} & \textbf{57.2609} & \textbf{57.2609}  \\

\cline{2-10}
   &SMOKE& C & & 1.6560 & 1.6251 & 1.6251 & 2.2562 & 2.2285 & 2.2285 \\
  &PointPillars& L  & Barrier  & 26.9185 & 26.7128 & 26.7128  & 34.5679 & 34.1097 & 34.1097    \\
  &MVXNet& C, L &  & \textbf{32.8250} & \textbf{32.8064} & \textbf{32.8064} &  \textbf{42.2126} & \textbf{42.1718} & \textbf{42.1718}  \\
\hline
\hline
 Sim &SMOKE& C       & & 0.5020& 0.4643 & 0.4643 & 1.0534 & 0.6718 & 0.6718 \\
  &PointPillars& L  & Road beacon  & 93.4512 & 95.2941 & 95.4048 & 93.4512 & 95.3508 & 95.4544   \\
   & MVXNet & C, L &  & \textbf{97.3106} & \textbf{97.3035} & \textbf{97.3106} &  \textbf{97.3106} & \textbf{97.3035} & \textbf{97.3106}  \\

\cline{2-10}
   &SMOKE& C & & 0.5282 & 0.5282 & 0.5282& 1.5635 &  1.5455 & 1.5455 \\
  &PointPillars& L  & Barrier  &\textbf{74.4662} & 74.3497 & 74.3272 & \textbf{74.4662} & 74.2039 & 74.1847    \\
  &MVXNet& C, L  &  & 74.1536 & \textbf{74.4607} & \textbf{74.4351} & 74.1610 & \textbf{74.4643} & \textbf{74.4388}  \\
\hline
\end{tabular}

\end{center}
\end{table*}

\subsection{Tracker-based Object Geo-localization} \textbf{Experiment Details}. We assess the object geo-localization module by evaluating two tracklet fusion strategies applied to the outputs from three 3D object detectors. In addition, the ground-truth 3D bounding boxes are used as inputs to object geo-localization module for reference evaluation. Specifically, we conduct experiments on 3 consecutive scenes from the RZDG-Real dataset and 8 consecutive scenes from 8 different towns from the RZDG-Sim dataset.

\textbf{Evaluation}.
The accuracy of the Roadwork Zone Detection and Geo-localization (RZDG) pipeline is presented in Table \ref{tab:geo_results}. For this evaluation, we adopt several well-established metrics, including Precision, Recall, and F1-score. Building on methodologies outlined in prior studies \cite{Wilson, Nassar}, we further refine our evaluation by employing the Haversine formula to accurately quantify geographic discrepancies between predicted and actual locations. A True Positive (TP) is defined as any predicted location that lies within a 1-meter threshold of its corresponding ground truth, with the distance calculated using the Haversine formula.

The Haversine distance ($d$) between predicted and ground-truth locations in the global coordinate system is calculated using the Haversine formula as follows:

\begin{equation} a = \sin^2\left(\frac{\Delta \phi}{2}\right) + \cos(\phi_1) \cdot \cos(\phi_2) \cdot \sin^2\left(\frac{\Delta \lambda}{2}\right) \end{equation} \begin{equation} c = 2 \cdot \operatorname{atan2}\left(\sqrt{a}, \sqrt{1-a}\right) \end{equation} \begin{equation} d = R \cdot c \end{equation}

Here, $\phi_1$ and $\phi_2$ denote the latitudes of the predicted and ground-truth locations, respectively. The longitudes $\lambda_1$ and $\lambda_2$ correspond to the predicted and ground-truth positions. $\Delta \phi$ represents the latitude difference ($\phi_1 - \phi_2$), and $\Delta \lambda$ represents the longitude difference ($\lambda_1 - \lambda_2$). The Earth's radius ($R$) is assumed to be 6,372,800 meters. The value $d$ indicates the spherical distance between the predicted and ground-truth positions on the Earth’s surface.

\begin{table}[ht]
\caption{Geo-localization Results on RZDG Dataset (LF=Last Frame, WA=Weighted Average). Results are shown with the best in each group in bold without GT.\label{tab:geo_results}}
\begin{center}
\begin{tabular}{p{0.5cm}|p{0.8cm}p{1cm}p{0.8cm}|p{0.8cm}p{0.8cm}p{0.8cm}}
\hline
&Strategies &Methods&H. thres.$^{\mathrm{a}}$ & Precision& Recall& F1-Score  \\
\hline
Real&WA  & GT $^{\mathrm{b}}$    & 1 & 0.719 & 0.787 & 0.749 \\
&WA &SMOKE                       & 1 & 0.467 & 0.491 & 0.478    \\
&WA &PointPillars                & 1 & 0.190 & 0.864 & 0.272    \\
&WA &MVXNet                      & 1 & 0.449 & 0.880 & 0.589    \\
\cline{2-7} 
&LF &GT $^{\mathrm{b}}$          & 1 & 0.844  & 0.954  & 0.892    \\
&LF &SMOKE                       & 1 & \textbf{0.565}  & 0.598  & 0.580    \\
&LF &PointPillars                & 1 & 0.415  & 0.845  & 0.525    \\
&LF &MVXNet                      & 1 & 0.452  & \textbf{0.898}  & \textbf{0.597}    \\
\cline{2-7}

\hline
\hline

Sim& WA  & GT$^{\mathrm{b}}$     & 1 & 0.769 & 0.769 & 0.769 \\
&WA  &SMOKE                      & 1 & 0.218 & 0.521 & 0.304 \\
&WA  & PointPillars              & 1 & 0.515 & \textbf{0.809} & 0.626 \\
&WA  & MVXNet                    & 1 & \textbf{0.615} & 0.745 & \textbf{0.665} \\
\cline{2-7} 
&LF & GT$^{\mathrm{b}}$          & 1 & 0.173  & 0.173  & 0.173    \\
&LF & SMOKE                      & 1 & 0.125  & 0.303  & 0.175    \\
&LF & PointPillars               & 1 & 0.133  & 0.205  & 0.160    \\
&LF & MVXNet                     & 1 & 0.162  & 0.211  & 0.180    \\
\cline{2-7}

\hline
\multicolumn{7}{l}{$^{\mathrm{a}}$Haversine threshold (meter)} \\
\multicolumn{7}{l}{$^{\mathrm{b}}$ground-truth 3D bounding box for reference} \\

\end{tabular}
\end{center}
\end{table}

\section{Image Semantic Segmentation Task}
Our dataset also supports an image semantic segmentation task and benchmarks three existing algorithms. This task can be used to detect the presence of roadwork zones.

\textbf{Experiment Details}. Three algorthms  \cite{deeplabv3p}, \cite{swin} and \cite{segformer} are trained and evaluated on the RZDG-Real dataset and the RZDG-Sim dataset, respectively. The RZDG-Real dataset is split into a training set and a validation set as per the ratio of 0.8:0.2, leading to 687 training samples and 172 validation samples. The RZDG-Sim dataset is split into a training set and a validation set as per the ratio of 0.8:0.2, leading to 6420 training samples and 802 validation samples. The experiments are performed on a computing platform equipped with a NVIDIA RTX 4080 GPU. All semantic segmentation algorithms are implemented in mmSegmentation \cite{mmsegmentation}.

\textbf{Evaluation}. Following the evaluation of semantic segmentation in previous work, such as \cite{deeplabv3p}, \cite{swin} and \cite{segformer}, we utilize the Intersection over Union (IoU) and Accuracy (Acc) for evaluation. Additionally, the mean IoU (mIoU) is calculated as the average of the IoUs across all classes. the mean Acc (mAcc) is the average of the accuracy of all classes. Table ~\ref{tab:segmentation_results} shows the performance of the three algorithms on two categories and the inference time of each algorithm of one frame utilizing one NVIDIA RTX 4080 GPU.

\begin{table}[h]
\caption{Image Semantic Segmentation Results on RZDG Dataset. Results are shown with the best in each group in bold.}
\label{tab:segmentation_results}
\centering
\begin{tabular}{p{0.4cm}|p{1.5cm}|p{0.5cm}|p{0.5cm}|p{0.5cm}|p{0.5cm}|p{0.5cm}|p{0.5cm}}
\hline
&Class & \multicolumn{2}{c|}{Deeplabv3+} & \multicolumn{2}{c|}{Swin} & \multicolumn{2}{c}{Segformer} \\
\cline{3-8} 
 & &IoU & Acc & IoU & Acc & IoU & Acc\\
\hline
Real& Barrier      & \textbf{93.37} & \textbf{96.84} & 92.53 & 96.39 & 88.34 & 92.64 \\
    & Road beacon  & \textbf{85.92} & \textbf{91.54} & 82.19 & 88.89 & 76.22 & 82.55 \\
\cline{2-8} 
&mIoU        & \multicolumn{2}{c|}{\textbf{92.98}} & \multicolumn{2}{c|}{91.44} & \multicolumn{2}{c}{87.99} \\
&mAcc        & \multicolumn{2}{c|}{\textbf{96.07}} & \multicolumn{2}{c|}{95.02} & \multicolumn{2}{c}{91.66}\\
&Infer.time  & \multicolumn{2}{c|}{128 ms} & \multicolumn{2}{c|}{135 ms} & \multicolumn{2}{c}{\textbf{80 ms}}\\
\hline
\hline
Sim& Barrier      & 89.38 & 94.29 & \textbf{89.84} & \textbf{94.76} & 77.68 & 85.57\\
   & Road beacon  & 91.63 & 96.06 & \textbf{91.85} & \textbf{96.35} & 85.35 & 92.54 \\
\cline{2-8} 
&mIoU        & \multicolumn{2}{c|}{83.52} & \multicolumn{2}{c|}{\textbf{83.98}}  & \multicolumn{2}{c}{71.87}\\
&mAcc        & \multicolumn{2}{c|}{\textbf{89.21}} & \multicolumn{2}{c|}{88.54} & \multicolumn{2}{c}{77.56}\\
&Infer.time  & \multicolumn{2}{c|}{101 ms} & \multicolumn{2}{c|}{108 ms} & \multicolumn{2}{c}{\textbf{63 ms}} \\
\hline
\end{tabular}
\end{table}

\section{Conclusion} To address the challenges posed by transient road work zones, which are typically not documented in high-definition (HD) maps, we introduce the Roadwork Zone Detection and Geo-localization (RZDG) dataset, featuring both real-world and simulated data. We also present a comprehensive Roadwork Zone Detection and Geo-localization pipeline that leverages multi-modal sensors, including camera, LiDAR, and GPS/IMU to predict the object's global position. Despite setting a stringent accuracy threshold with a maximum error of one meter between predicted and actual positions, our methodology consistently delivers superior results.

\section*{ACKNOWLEDGMENT}
We would like to thank many people (in alphabetical order) who contributed to the development of the simulation environment, sensor car and the construction of the dataset: Shiva Agrawal, Prithviraj Basu, Marcel Kettelgerdes, Raksha Padiyar, and many more.


\bibliographystyle{IEEEtran}
\bibliography{ref}

\begin{thebibliography}{10}
\providecommand{\url}[1]{#1}
\csname url@samestyle\endcsname
\providecommand{\newblock}{\relax}
\providecommand{\bibinfo}[2]{#2}
\providecommand{\BIBentrySTDinterwordspacing}{\spaceskip=0pt\relax}
\providecommand{\BIBentryALTinterwordstretchfactor}{4}
\providecommand{\BIBentryALTinterwordspacing}{\spaceskip=\fontdimen2\font plus
\BIBentryALTinterwordstretchfactor\fontdimen3\font minus
  \fontdimen4\font\relax}
\providecommand{\BIBforeignlanguage}[2]{{%
\expandafter\ifx\csname l@#1\endcsname\relax
\typeout{** WARNING: IEEEtran.bst: No hyphenation pattern has been}%
\typeout{** loaded for the language `#1'. Using the pattern for}%
\typeout{** the default language instead.}%
\else
\language=\csname l@#1\endcsname
\fi
#2}}
\providecommand{\BIBdecl}{\relax}
\BIBdecl

\bibitem{garcia2022ildm}
M.~Garc{\'\i}a, I.~Urbieta, M.~Nieto, J.~Gonz{\'a}lez~de Mendibil, and
  O.~Otaegui, ``ildm: An interoperable graph-based local dynamic map,''
  \emph{Vehicles}, vol.~4, no.~1, pp. 42--59, 2022.

\bibitem{etsi2009intelligent}
T.~ETSI, ``Intelligent transport systems (its); vehicular communications; basic
  set of applications; definitions,'' \emph{Tech. Rep. ETSI TR 102 6382009},
  2009.

\bibitem{Sundharam}
V.~Sundharam, A.~Sarkar, A.~Svetovidov, J.~S. Hickman, and A.~L. Abbott,
  ``Characterization, detection, and segmentation of work-zone scenes from
  naturalistic driving data,'' \emph{Transportation research record}, vol.
  2677, no.~3, pp. 490--504, 2023.

\bibitem{Oh}
D.~Oh, K.~Kang, S.~Seo, J.~Xiao, K.~Jang, K.~Kim, H.~Park, and J.~Won,
  ``Low-cost object detection models for traffic control devices through domain
  adaption of geographical regions,'' \emph{Remote Sensing}, vol.~15, no.~10,
  p. 2584, 2023.

\bibitem{Katsamenis}
I.~Katsamenis, E.~E. Karolou, A.~Davradou, E.~Protopapadakis, A.~Doulamis,
  N.~Doulamis, and D.~Kalogeras, ``Tracon: A novel dataset for real-time
  traffic cones detection using deep learning,'' in \emph{Novel \& Intelligent
  Digital Systems Conferences}.\hskip 1em plus 0.5em minus 0.4em\relax
  Springer, 2022, pp. 382--391.

\bibitem{Shi}
W.~Shi and R.~R. Rajkumar, ``Work zone detection for autonomous vehicles,'' in
  \emph{2021 IEEE International Intelligent Transportation Systems Conference
  (ITSC)}.\hskip 1em plus 0.5em minus 0.4em\relax IEEE, 2021, pp. 1585--1591.

\bibitem{weng2020ab3dmot}
X.~Weng, J.~Wang, D.~Held, and K.~Kitani, ``Ab3dmot: A baseline for 3d
  multi-object tracking and new evaluation metrics,'' \emph{arXiv preprint
  arXiv:2008.08063}, 2020.

\bibitem{Shahin}
F.~Shahin, W.~Elias, and T.~Toledo, ``Effects of highway work zone temporary
  countermeasures,'' \emph{European Transport Research Review}, vol.~15, no.~1,
  p.~20, 2023.

\bibitem{deeplabv3p}
L.-C. Chen, Y.~Zhu, G.~Papandreou, F.~Schroff, and H.~Adam, ``Encoder-decoder
  with atrous separable convolution for semantic image segmentation,'' in
  \emph{Proceedings of the European conference on computer vision (ECCV)},
  2018, pp. 801--818.

\bibitem{swin}
Z.~Liu, Y.~Lin, Y.~Cao, H.~Hu, Y.~Wei, Z.~Zhang, S.~Lin, and B.~Guo, ``Swin
  transformer: Hierarchical vision transformer using shifted windows,'' in
  \emph{Proceedings of the IEEE/CVF international conference on computer
  vision}, 2021, pp. 10\,012--10\,022.

\bibitem{segformer}
E.~Xie, W.~Wang, Z.~Yu, A.~Anandkumar, J.~M. Alvarez, and P.~Luo, ``Segformer:
  Simple and efficient design for semantic segmentation with transformers,''
  \emph{Advances in neural information processing systems}, vol.~34, pp.
  12\,077--12\,090, 2021.

\bibitem{li2019gs3d}
B.~Li, W.~Ouyang, L.~Sheng, X.~Zeng, and X.~Wang, ``Gs3d: An efficient 3d
  object detection framework for autonomous driving,'' in \emph{Proceedings of
  the IEEE/CVF conference on computer vision and pattern recognition}, 2019,
  pp. 1019--1028.

\bibitem{cai2020monocular}
Y.~Cai, B.~Li, Z.~Jiao, H.~Li, X.~Zeng, and X.~Wang, ``Monocular 3d object
  detection with decoupled structured polygon estimation and height-guided
  depth estimation,'' in \emph{Proceedings of the AAAI Conference on Artificial
  Intelligence}, vol.~34, no.~07, 2020, pp. 10\,478--10\,485.

\bibitem{Mono3d}
X.~Chen, K.~Kundu, Z.~Zhang, H.~Ma, S.~Fidler, and R.~Urtasun, ``Monocular 3d
  object detection for autonomous driving,'' in \emph{Proceedings of the IEEE
  conference on computer vision and pattern recognition}, 2016, pp. 2147--2156.

\bibitem{liu2021ground}
Y.~Liu, Y.~Yixuan, and M.~Liu, ``Ground-aware monocular 3d object detection for
  autonomous driving,'' \emph{IEEE Robotics and Automation Letters}, vol.~6,
  no.~2, pp. 919--926, 2021.

\bibitem{SMOKE}
Z.~Liu, Z.~Wu, and R.~T{\'o}th, ``Smoke: Single-stage monocular 3d object
  detection via keypoint estimation,'' in \emph{Proceedings of the IEEE/CVF
  conference on computer vision and pattern recognition workshops}, 2020, pp.
  996--997.

\bibitem{MonoFlex}
Y.~Zhang, J.~Lu, and J.~Zhou, ``Objects are different: Flexible monocular 3d
  object detection,'' in \emph{Proceedings of the IEEE/CVF Conference on
  Computer Vision and Pattern Recognition}, 2021, pp. 3289--3298.

\bibitem{PointNet}
R.~Q. Charles, H.~Su, M.~Kaichun, and L.~J. Guibas, ``Pointnet: Deep learning
  on point sets for 3d classification and segmentation,'' in \emph{2017 IEEE
  Conference on Computer Vision and Pattern Recognition (CVPR)}, 2017, pp.
  77--85.

\bibitem{DGCNN}
\BIBentryALTinterwordspacing
Y.~Wang, Y.~Sun, Z.~Liu, S.~E. Sarma, M.~M. Bronstein, and J.~M. Solomon,
  ``Dynamic graph cnn for learning on point clouds,'' \emph{ACM Trans. Graph.},
  vol.~38, no.~5, Oct. 2019. [Online]. Available:
  \url{https://doi.org/10.1145/3326362}
\BIBentrySTDinterwordspacing

\bibitem{KPConv}
H.~Thomas, C.~R. Qi, J.-E. Deschaud, B.~Marcotegui, F.~Goulette, and L.~J.
  Guibas, ``Kpconv: Flexible and deformable convolution for point clouds,'' in
  \emph{Proceedings of the IEEE/CVF International Conference on Computer Vision
  (ICCV)}, October 2019.

\bibitem{VoxelNet}
Y.~Zhou and O.~Tuzel, ``Voxelnet: End-to-end learning for point cloud based 3d
  object detection,'' in \emph{Proceedings of the IEEE Conference on Computer
  Vision and Pattern Recognition (CVPR)}, June 2018.

\bibitem{SECOND}
\BIBentryALTinterwordspacing
L.~Zhang, H.~Meng, Y.~Yan, and X.~Xu, ``Transformer-based global pointpillars
  3d object detection method,'' \emph{Electronics}, vol.~12, no.~14, p. 3092,
  2023. [Online]. Available: \url{https://doi.org/10.3390/electronics12143092}
\BIBentrySTDinterwordspacing

\bibitem{PV-RCNN}
S.~Shi, C.~Guo, L.~Jiang, Z.~Wang, J.~Shi, X.~Wang, and H.~Li, ``Pv-rcnn:
  Point-voxel feature set abstraction for 3d object detection,'' in
  \emph{Proceedings of the IEEE/CVF Conference on Computer Vision and Pattern
  Recognition (CVPR)}, June 2020.

\bibitem{PointPillars}
A.~H. Lang, S.~Vora, H.~Caesar, L.~Zhou, J.~Yang, and O.~Beijbom,
  ``Pointpillars: Fast encoders for object detection from point clouds,'' in
  \emph{Proceedings of the IEEE/CVF Conference on Computer Vision and Pattern
  Recognition (CVPR)}, June 2019.

\bibitem{limits}
X.~Zhao, P.~Sun, Z.~Xu, H.~Min, and H.~Yu, ``Fusion of 3d lidar and camera data
  for object detection in autonomous vehicle applications,'' \emph{IEEE Sensors
  Journal}, vol.~20, no.~9, pp. 4901--4913, 2020.

\bibitem{MVXNet}
V.~A. Sindagi, Y.~Zhou, and O.~Tuzel, ``Mvx-net: Multimodal voxelnet for 3d
  object detection,'' in \emph{2019 International Conference on Robotics and
  Automation (ICRA)}.\hskip 1em plus 0.5em minus 0.4em\relax IEEE, 2019, pp.
  7276--7282.

\bibitem{3DDualFusion}
Y.~Kim, K.~Park, M.~Kim, D.~Kum, and J.~W. Choi, ``3d dual-fusion: Dual-domain
  dual-query camera-lidar fusion for 3d object detection,'' \emph{arXiv
  preprint arXiv:2211.13529}, 2022.

\bibitem{Wilson_Zhang}
D.~Wilson, X.~Zhang, W.~Sultani, and S.~Wshah, ``Image and object
  geo-localization,'' \emph{International Journal of Computer Vision}, vol.
  132, no.~4, pp. 1350--1392, 2024.

\bibitem{govada2020road}
K.~A. Govada, H.~P. Jonnalagadda, P.~Kapavarapu, S.~Alavala, and K.~S. Vani,
  ``Road deformation detection,'' in \emph{2020 7th International Conference on
  Smart Structures and Systems (ICSSS)}.\hskip 1em plus 0.5em minus 0.4em\relax
  IEEE, 2020, pp. 1--5.

\bibitem{Wilson}
D.~Wilson, T.~Alshaabi, C.~Van~Oort, X.~Zhang, J.~Nelson, and S.~Wshah,
  ``Object tracking and geo-localization from street images,'' \emph{Remote
  Sensing}, vol.~14, no.~11, p. 2575, 2022.

\bibitem{Chaabane}
M.~Chaabane, L.~Gueguen, A.~Trabelsi, R.~Beveridge, and S.~O'Hara, ``End-to-end
  learning improves static object geo-localization from video,'' in
  \emph{Proceedings of the IEEE/CVF Winter Conference on Applications of
  Computer Vision}, 2021, pp. 2063--2072.

\bibitem{Krylov}
V.~A. Krylov, E.~Kenny, and R.~Dahyot, ``Automatic discovery and geotagging of
  objects from street view imagery,'' \emph{Remote Sensing}, vol.~10, no.~5, p.
  661, 2018.

\bibitem{Namazi}
E.~Namazi, R.~Mester, C.~Lu, and J.~Li, ``Geolocation estimation of target
  vehicles using image processing and geometric computation,''
  \emph{Neurocomputing}, vol. 499, pp. 35--46, 2022.

\bibitem{kitti}
A.~Geiger, P.~Lenz, C.~Stiller, and R.~Urtasun, ``Vision meets robotics: The
  kitti dataset,'' \emph{The International Journal of Robotics Research},
  vol.~32, no.~11, pp. 1231--1237, 2013.

\bibitem{macenski2022robot}
S.~Macenski, T.~Foote, B.~Gerkey, C.~Lalancette, and W.~Woodall, ``Robot
  operating system 2: Design, architecture, and uses in the wild,''
  \emph{Science robotics}, vol.~7, no.~66, p. eabm6074, 2022.

\bibitem{bibliocad_audi_a6}
\BIBentryALTinterwordspacing
{Bibliocad}, ``Audi a6 avant.'' [Online]. Available:
  \url{https://www.bibliocad.com/de/library/audi-a6-avant-_12984/}
\BIBentrySTDinterwordspacing

\bibitem{OpenStreetMap}
{OpenStreetMap contributors}, ``{Planet dump retrieved from
  https://planet.osm.org },'' \url{ https://www.openstreetmap.org }, 2017.

\bibitem{zhang2000flexible}
Z.~Zhang, ``A flexible new technique for camera calibration,'' \emph{IEEE
  Transactions on pattern analysis and machine intelligence}, vol.~22, no.~11,
  pp. 1330--1334, 2000.

\bibitem{luo2023calib}
Z.~Luo, G.~Yan, and Y.~Li, ``Calib-anything: Zero-training lidar-camera
  extrinsic calibration method using segment anything,'' \emph{arXiv preprint
  arXiv:2306.02656}, 2023.

\bibitem{10533619}
R.~Varghese and S.~M., ``Yolov8: A novel object detection algorithm with
  enhanced performance and robustness,'' in \emph{2024 International Conference
  on Advances in Data Engineering and Intelligent Computing Systems (ADICS)},
  2024, pp. 1--6.

\bibitem{li2020sustech}
E.~Li, S.~Wang, C.~Li, D.~Li, X.~Wu, and Q.~Hao, ``Sustech points: A portable
  3d point cloud interactive annotation platform system,'' in \emph{2020 IEEE
  Intelligent Vehicles Symposium (IV)}.\hskip 1em plus 0.5em minus 0.4em\relax
  IEEE, 2020, pp. 1108--1115.

\bibitem{5483185}
A.~Torralba, B.~C. Russell, and J.~Yuen, ``Labelme: Online image annotation and
  applications,'' \emph{Proceedings of the IEEE}, vol.~98, no.~8, pp.
  1467--1484, 2010.

\bibitem{dosovitskiy2017carla}
A.~Dosovitskiy, G.~Ros, F.~Codevilla, A.~Lopez, and V.~Koltun, ``Carla: An open
  urban driving simulator,'' in \emph{Conference on robot learning}.\hskip 1em
  plus 0.5em minus 0.4em\relax PMLR, 2017, pp. 1--16.

\bibitem{schonborn2022rsa}
H.~D. Sch{\"o}nborn and W.~Schulte, \emph{RSA Handbuch: Richtlinien f{\"u}r die
  verkehrsrechtliche Sicherung von Arbeitsstellen an Stra{\ss}en RSA 21,
  Ausgabe 2021: Handbuch und Kommentar}, 2022.

\bibitem{heidenreich1976distance}
C.~Heidenreich, ``Distance, latitude, longitude, and navigation,''
  \emph{Cartographica: The International Journal for Geographic Information and
  Geovisualization}, vol.~13, no.~2, pp. 42--76, 1976.

\bibitem{mmdet3d2020}
M.~Contributors, ``{MMDetection3D: OpenMMLab} next-generation platform for
  general {3D} object detection,''
  \url{https://github.com/open-mmlab/mmdetection3d}, 2020.

\bibitem{Nassar}
A.~S. Nassar, S.~Lef{\`e}vre, and J.~D. Wegner, ``Simultaneous multi-view
  instance detection with learned geometric soft-constraints,'' in
  \emph{Proceedings of the IEEE/CVF international conference on computer
  vision}, 2019, pp. 6559--6568.

\bibitem{mmsegmentation}
K.~Chen, J.~Wang, J.~Pang, Y.~Cao, Y.~Xiong, X.~Li, S.~Sun, W.~Feng, Z.~Liu,
  J.~Xu \emph{et~al.}, ``Mmdetection: Open mmlab detection toolbox and
  benchmark,'' \emph{arXiv preprint arXiv:1906.07155}, 2019.

\end{thebibliography}

\end{document}